\documentclass{article}
\usepackage{spconf,amsmath,graphicx}
\usepackage{color}
\usepackage{graphicx}
\usepackage{subfigure}

\usepackage{titlesec}
\titlespacing\section{0pt}{5pt minus 2pt}{2pt plus 2pt}
\titlespacing\subsection{0pt}{2pt minus 2pt}{2pt plus 2pt}
\titlespacing\subsubsection{0pt}{2pt minus 2pt}{2pt plus 2pt}

\title{MEmoBERT: Pre-training Model with Prompt-based Learning for Multimodal Emotion Recognition}

\name{Jinming Zhao$^{1,2}$, Ruichen Li$^{1}$, Qin Jin$^{1}$*\thanks{*Corresponding author.}, Xinchao  Wang$^{2}$, Haizhou Li$^{2}$}
\address{$^{1}$ School of Information, Renmin University of China \\
			    $^{2}$ Electrical and Computer Engineering, National University of Singapore}
\begin{document}
%
\maketitle
\begin{abstract}
Multimodal emotion recognition study is hindered by the lack of labelled corpora in terms of scale and diversity, due to the high annotation cost and label ambiguity.
In this paper, we propose a pre-training model \textbf{MEmoBERT} for multimodal emotion recognition, which learns multimodal joint representations through self-supervised learning from large-scale unlabeled video data that come in sheer volume. 
Furthermore, unlike the conventional ``pre-train, finetune'' paradigm, we propose a prompt-based method that reformulates the downstream emotion classification task as a masked text prediction one, bringing the downstream task closer to the pre-training.
Extensive experiments on two benchmark datasets, IEMOCAP and MSP-IMPROV, show that our proposed MEmoBERT significantly enhances emotion recognition performance.

\end{abstract}
\begin{keywords}
Emotion Recognition, Multimodal, Pre-training, Prompt
\end{keywords}


\begin{figure*}[htp!]
\centering
\includegraphics[scale=0.83]{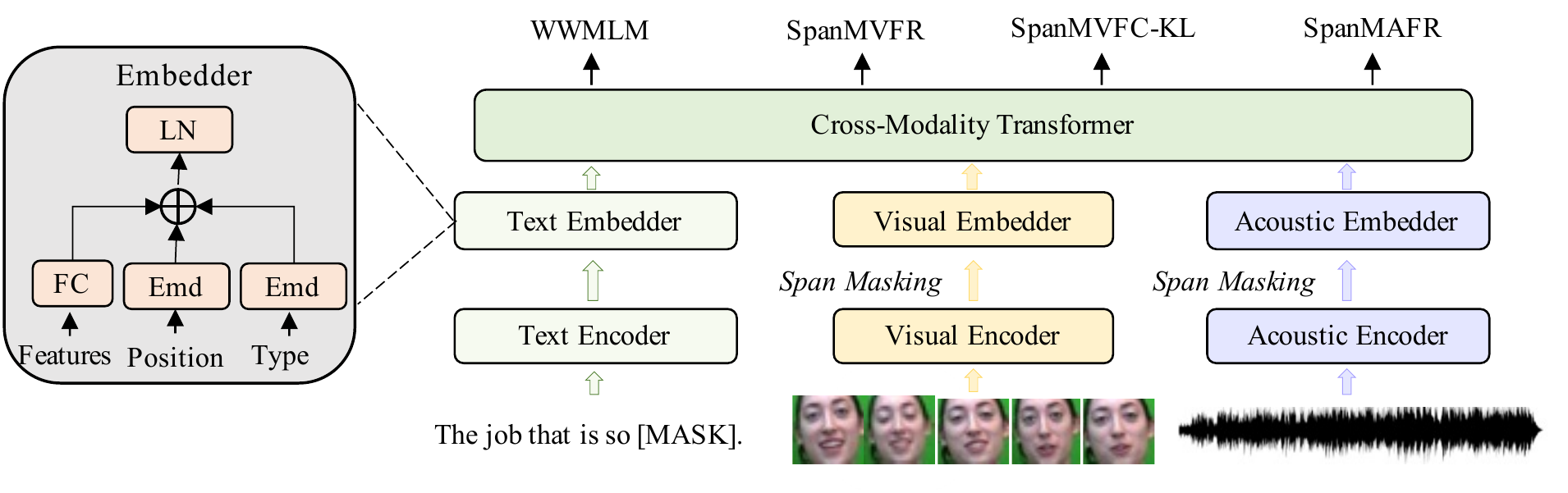}
\vspace{-2pt}
\caption{Overview of the proposed MEmoBERT model consisting of three modality-specific Encoders, three modality-specific Embedders and a multi-layer Cross Modality Transformer, learned through four pre-training tasks. ``FC'' and ``Emd'' in Embedder are a fully-connected layer and a embedding layer respectively. ``Type'' in Embedder refers different modalities, for example, using 0 as text modality type. Different modality Embedders have same structure but individual parameters.}
\label{fig:model}
\vspace{-9pt}
\end{figure*}

\section{Introduction}
\label{sec:intro}

Automatic Multimodal Emotion Recognition aims to interpret human emotions through multiple modalities such as text, audio and visual, that is an enabling technology for many applications including 
human-computer interactions \cite{Fragopanagos2002Emotion}. 
Many previous works have explored multimodal fusion strategies. 
For example, 
MFN \cite{zadeh2018memory} and MARN \cite{zadeh2018multi} are attention- and memory-based approaches applicable to sequences aligned at the word-level.
MulT \cite{tsai2019multimodal} is a transformer-based framework, which handles both non-aligned multimodal sequences and long-range dependencies.

In recent years, various pre-trained models via self-supervised learning on large-scale unlabeled data have achieved  promising results.
The pre-trained language representation models, such as BERT \cite{devlin2019bert}, ELMo \cite{peters2018deepelmo} and GPT \cite{radford2019GPT}, have attracted much attention and are widely adopted.
Inspired by the success of self-supervised pre-training in textual modality, many visual+language cross-modality pre-trained models are proposed \cite{bugliarello2021multimodalsurvey, su2019vlbert, chen2020uniter} and have achieved new state-of-the-art performances on various tasks, such as Image-Text Retrieval, Visual Question Answering etc. 
Such models, including UNITER \cite{chen2020uniter} and LXMERT \cite{tan2019lxmert}, typically follow a single- or dual-stream multi-layer Transformer architecture and optimize through several pre-training tasks to learn joint multimodal representations, such as Masked Language Modeling (MLM), Masked Region Modeling (MRM), Image-Text Matching (ITM).
VideoBERT \cite{sun2019videobert}, ActBERT \cite{zhu2020actbert} and HERO \cite{li2020hero} extend to the video+language representation learning for the video-related tasks, such as Video-Text Retrieval, Video Action Classification.
Furthermore, VATT \cite{akbari2021vatt} considers three modalities (text, visual and audio) with a modality-agnostic Transformer and uses multimodal contrastive learning to learn multimodal representations.

Prompt-based learning \cite{liu2021prompt}, on the other hand, has achieved great success and has become a new learning paradigm in NLP. 
Compared to the ``pre-train, finetune'' paradigm, which adapts pre-trained models to downstream tasks via objective engineering, the ``pre-train, prompt and predict'' paradigm reformulates the downstream tasks to resemble the masked language modeling task optimized in the original pre-training with the help of a textual prompt. This paradigm brings the downstream tasks closer to the pre-training tasks, which can retain more learned knowledge during pre-training.  It outperforms the ``pre-train, finetune'' paradigm, especially under low-resource conditions, and has achieved promising results in many NLP tasks, such as Question Answering \cite{petroni2019lama, zhong2021adaptingmeta}, Text Classification \cite{schick2020fewpet, schick2021exploitingPET}.

Motivated by the above studies, we propose a multimodal transformer-based pre-training model, MEmoBERT, to learn joint multimodal representations for emotion recognition. It is trained through self-supervised learning based on a large-scale unlabeled video dataset comprising more than 300 movies. We design four efficient self-supervised pre-training tasks to learn joint multimodal emotional representations, including Whole Word Masked Language Modeling (WWMLM), Span Masked Visual Frame Modeling (SpanMVFM), Span Masked Visual Frame Classification with KL-divergence (SpanMVFC-KL), Span Masked Acoustic Frame Modeling (SpanMAFM).
Furthermore, we explore a prompt-based learning method to efficiently adapt the pre-trained model to multimodel emotion recognition. To the best of our knowledge, the proposed MEmoBERT is the first multimodal pre-training model in the area of multimodal emotion recognition, and it is also the first attempt to adopt prompt-based learning for this area.
We carry out experiments on two benchmark datasets, IEMOCAP and MSP-IMPROV, to evaluate our pre-trained MEmoBERT.
We specifically compare the proposed prompt-based learning method with the traditional ``pre-train, finetune'' method on full- and part-training data.
The experiments results demonstrate that our MEmoBERT yields significant improvement and the prompt-based learning method further improves the performance.

The main contributions of this work include,
1) We proposed a multimodal transformer-based pre-trained model, MEmoBERT, under self-supervised learning on large-scale unlabeled movies dataset for multimodel emotion recognition.
2) We propose a prompt-based learning method that better adapts the pre-trained MEmoBERT to downstream multimodal emotion recognition tasks.
3) Our proposed model achieves a new state-of-the-art performance on both IEMOCAP and MSP multimodal emotion recognition benchmark datasets.

\section{Method}
\label{sec:method}
Fig.~\ref{fig:model} illustrates the overall model framework of our proposed MEmoBERT and its learning process during pre-training.
MEmoBERT consists of three independent modality Encoders to generate modality-specific token- or frame-level raw features for the textual, visual, and acoustic modalities, and three Embedders to generate embeddings based on corresponding raw features respectively. Specifically, the embedding layer in BERT \cite{devlin2019bert} is adopted as the Text Encoder. The Visual Encoder is a pre-trained facial expression model that generates the visual expression features based on the speaker faces. The Acoustic Encoder is a pre-trained speech model that generates the acoustic features based on the audio waveform. The final embedding for each modality is obtained via its modality Embedder which sums up the raw features, position embedding and type embedding, and then gets normalized via Layer Norm. Please note that the parameters of Acoustic Encoder and Visual Encoder are fixed during pre-training. A cross-modality transformer in MEmoBERT then learns cross-modality contextualized representation based on the embeddings from different modalities.

We design four efficient pre-training tasks to optimize MEmoBERT in the pre-training phase to learn joint multimodal emotional representations. Once the model is well pre-trained,
we adopt the prompt-based learning method to adapt it to downstream tasks.

\subsection{Cross Modality Transformer}
The cross-modality transformer adopts the most established Transformer architecture \cite{devlin2019bert} and extends it to three modalities (text, visual and audio) for multimodal pre-training.
We follow the modality-agnostic strategy \cite{chen2020uniter, akbari2021vatt}, that is, a single backbone Transformer is applied to any of the modalities. During pre-training, the modality-specific embeddings are fed into the multi-layer Transformer to learn high-level cross-modality contextualized representations across different modalities.

\subsection{Pre-training Tasks}
\label{sec:pre-tasks}
We design four pre-training tasks including text, visual and audio modality-related tasks to enhance the cross-modality interaction and to learn joint multimodal emotional representations. In all following pre-training tasks, we adopt the conditional masking strategy instead of applying joint random masking to all modalities. It only masks one modality and keeps other modalities intact in corresponding tasks, which can learn the better latent alignment across three modalities and enables the model to learn better joint multimodal representations effectively \cite{chen2020uniter}. For example, as shown in Fig.~\ref{fig:model}, in the case where the word ``cool'' is masked (WWMLM), our model should be able to infer the masked word based on the surrounding text, facial expressions and acoustic signal. While in the case where the facial expression features of several smiley faces are masked (SpanMVFR/SpanMVFC-KL), our model should be able to infer the facial expressions features / emotion distribution of the masked visual frames based on the surrounding facial expressions, the word ``cool'', and the voice tone.

\noindent \textbf{Whole Word Masked Language Modeling (WWMLM)} learns to predict the masked whole word conditioned on the visual and the acoustic modalities. The whole word masking strategy that masks whole word rather than sub-tokens can better capture the accurate semantics~\cite{cui2019wwm}. For example, masking partial WordPiece tokens of a word may lead to completely opposite emotional semantics of the whole word, especially for words with the prefixes (e.g. ``un-'', ``im-'', ``op-'') and the suffixes (e.g ``-less'' ).

\noindent \textbf{Span Masked Acoustic Frame Regression (SpanMAFR)} learns to reconstruct the masked acoustic frame features conditioned on the textual and visual modalities \cite{chen2020uniter}. We apply L2 regression as the objective function to minimize the reconstruction error between the predicted and ground-truth frames. Furthermore, inspired by the span masking strategy \cite{liu2020mockingjay, jiang2021further} that aims to avoid the model exploiting the local smoothness of acoustic frames, we apply the span masking strategy that masks consecutive frames to zero. It can ensure the model to capture global emotional expression rather than local information. 

\noindent \textbf{Span Masked Visual Frame Regression (SpanMVFR)} learns to reconstruct the masked visual facial expression features conditioned on the textual and acoustic modalities \cite{chen2020uniter}. Due to the similarity of consecutive visual frames, similar to that in the acoustic modality, we also apply the span masking strategy for the visual modality. 

\noindent \textbf{Span Masked Visual Frame Classification with KL-divergence (SpanMVFC-KL)} learns to predict the facial expression class (such as happy, sad, anger) for each masked visual frame conditioned on the textual and acoustic modalities. We first feed the Transformer output of the masked frame into a fully connected layer to predict the emotion distribution of $K$ facial expression classes. Finally, we use the KL-divergence objective function to optimize the predicted emotion distributions with respect to the ground-truth emotion distributions which is produced by a pre-trained facial expression recognition model (Sec.~\ref{subsec:encs}). 

\subsection{Prompt-based Emotion Classification}
\begin{figure}[tp!]
\centering
\includegraphics[scale=0.70]{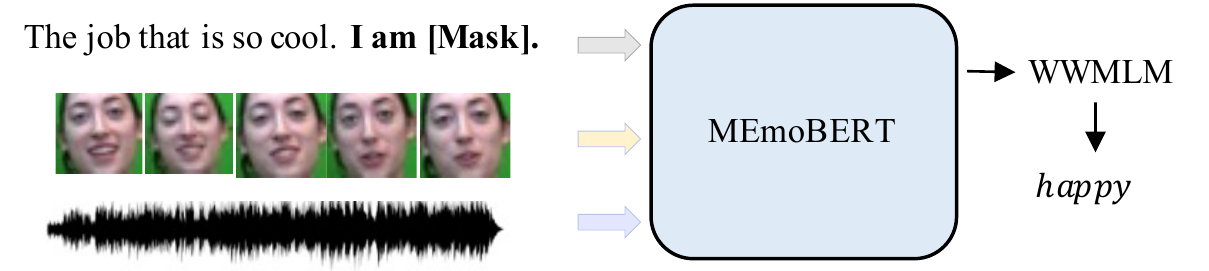}
\caption{Prompt-Prediction based on pre-trained MEmoBERT for downstream tasks.}
\label{fig:prompt}
\end{figure}
Fig.~\ref{fig:prompt} illustrates the ``prompt, predict'' paradigm. 
Given a prompt-based multimodal input ``[X] I am [MASK]. [V] [A]'' where the $[X], [V], [A]$ are text, visual and acoustic inputs of the video respectively, the classification problem is thus reformulated to predict the ``[MASK]'' as an emotion category word  (such as happy, sad, anger) with the help of a textual prompt ``I am [MASK].''. It is very similar to the whole word masked language modeling task in the pre-training phase.
\section{Experiments}
\label{sec:exprs}

\subsection{Pre-training Dataset}
Learning a pre-trained model for multimodal emotion recognition requires large-scale multimodal emotional data. We collect 351 movies and TV series that should belong to these categories, such as family, romance, soap opera, which have rich and natural emotional expressions. We extract the speakers' faces, audio, and corresponding subtitles of each utterance and filter out the empty utterances. In the end, we build a pre-training dataset containing about 180k utterances with three modalities.

\subsection{Benchmark Datasets}
\label{exp:dataset}
We evaluate our proposed MEmoBERT model on two benchmark multimodal emotion recognition datasets, including IEMOCAP \cite{IEMOCAP} and MSP-IMPROV \cite{busso2016msp}. We follow the emotional label processing in \cite{zhaomissing} to form the four-class emotion recognition setup. The statistics of the datasets are shown in Table.~\ref{tab:iemocap}.

\begin{table}[h]
    \centering
    \scalebox{0.9}{
    \begin{tabular}{cccccc}
         \hline
          dataset & Happy & Anger & Sadness & Neutral & Total \\
         \hline 
          IECMOAP & 1636 & 1103 & 1084 & 1708 & 5531 \\
         \hline
          MSP-IMPROV & 999 & 460 & 627 & 1733 & 3819 \\
          \hline
    \end{tabular}}
    \caption{A summary of the benchmark datasets.}
    \label{tab:iemocap}
\vspace{-3pt}
\end{table}

\subsection{Implementation Details}
\subsubsection{Modality Raw Feature Extraction}
\label{subsec:encs}
\textbf{Acoustic:} We extract the frame-level acoustic features from a pre-trained Wav2Vec2.0 model \cite{baevski2020wav2vec}. We sub-sample the frame-level features by average pooling every 3 frames.

 \textbf{Visual:} We first design an active speaker detection strategy based on the consistency of voice activation and mouth movement to get the speaker's faces. We then extract the face-level features and emotional probability distributions of the speaker's faces from a pre-trained DenseNet model \cite{huang2017densely} trained on a facial expression corpus, FER+ \cite{barsoum2016fer}.

\subsubsection{Experiment Setups}
During MEmoBERT pre-training, we first initialize its weights from a text pre-trained BERT checkpoint\footnote{https://huggingface.co/bert-base-uncased}. Specifically, MEmoBERT uses the same backbone architecture as BERT. 
For text modality, we follow the masking strategy used in BERT \cite{devlin2019bert}.
For the visual and acoustic masking strategy, we follow Mockingjay \cite{liu2020mockingjay} and set the consecutive masking number as 3.
We use AdamW optimizer with initial learning rate of 5e-5 over maximum 40K steps. 
The batch size is 640.

For experiments of the downstream tasks, we use the 10-fold and 12-fold speaker-independent cross-validation to evaluate the models on IEMOCAP and MSP-IMPROV respectively. In each fold, we use one speaker for testing and the remaining speakers for training. We use the weighted accuracy (WA) and unweighted average recall (UAR) as the evaluation metrics.
We run three times for each experiment and report the average performance.
We set the initial learning rate as 5e-5 and 3e-5 for experiments on full- and part-training data respectively over maximum 15 epochs. The batch size is 32.

In order to verify the effectiveness of our model framework and the prompt-based learning method, we specifically define four experiment settings: 1) ``Direct'' denotes that we directly train the MEmoBERT followed by a new emotion classifier for downstream tasks from scratch. 2) ``BERT+Direct'' denotes that we directly finetune the MEmoBERT followed by a new emotion classifier for downstream tasks, in which the MEmoBERT is initialized by a pre-trained text BERT.
3) ``Pretrain+Finetune'' denotes that we finetune the pre-trained MEmoBERT followed by a new emotion classifier for downstream tasks. 4) ``Pretrain+Prompt'' denotes that we adopt the prompt-based learning method based on the pre-trained MEmoBERT without introducing any additional parameters for downstream tasks.

\subsection{Experiments Results}
Table~\ref{tab:main_expr} presents the multimodal emotion recognition results on the two benchmark datasets, IEMOCAP and MSP-IMPROV. 
Compared to other state-of-the-art models without pre-training in the first block, ``BERT+Direct'' achieves superior performance on both datasets, which demonstrates that the pre-trained BERT language modal can benefit the multimodal emotion recognition. ``Pretrain+Finetune'' based on our pre-trained MEmoBERT achieves significant improvement compared to ``BERT+Direct''. Furthermore, ``Pretrain+Prompt'' with prompt-based learning over our pre-trained MEmoBERT can bring additional improvement.

\begin{table}[htp]
\centering
\scalebox{0.76}{
    \begin{tabular}{c|cc|cc}
    \hline
    & \multicolumn{2}{c|}{IEMOCAP}  & \multicolumn{2}{c}{MSP-IMPROV}  \\
                                & WA        &    UAR        &WA          & UAR   \\ 
    \hline
     cLSTM-MMA \cite{pan2020mmma}   & 73.94\%  &  --       &  --      & --  \\
     SSMM \cite{ajun2020}           & 75.60\%  & 74.50\%   &  --      & --  \\
     MMIN \cite{zhaomissing}        & --      &  78.12\%   &  --      & 68.55\%  \\
    \hline
     Direct                     & 74.64\%  &  75.76\%     &  67.17\%     & 65.57\%  \\
     BERT+Direct                & 77.98\%  &  78.98\%     &  70.08\%     & 69.67\%  \\
     Pretrain+Finetune          & 79.63\%  &  80.61\% (\textcolor{blue}{+1.6})   &  71.77\% & 71.35\% (\textcolor{blue}{+1.7}) \\
     Pretrain+Prompt            & 80.01\%  &  81.09\% (\textcolor{blue}{+2.1})   &  72.36\% & 72.22\% (\textcolor{blue}{+2.5})  \\
    \hline
    \end{tabular}}
\caption{Multimodal emotion recognition performance comparison on IEMOCAP and MSP-IMPROV. The numbers in blue refer to the improvements compared to ``BERT+Direct''.}
\label{tab:main_expr}
\vspace{-3pt}
\end{table}

\subsection{Ablation Study}

\textbf{Ablation of the pre-training tasks}
We first investigate the impact of different pre-training tasks on the performance of MEmoBERT.
As shown in Table~\ref{tab:ablation}, all the pre-training tasks, including the span masking and whole word masking strategies, and the visual and acoustic related pre-training tasks, are beneficial for pre-training the MEmoBERT.

\begin{table}[htp]
\centering
\scalebox{0.85}{
    \begin{tabular}{c|cc|cc}
    \hline
    & \multicolumn{2}{c|}{IEMOCAP}  & \multicolumn{2}{c}{MSP-IMPROV}  \\
                            & WA        &    UAR        &WA          & UAR   \\ 
    \hline
     Pretrain+Prompt        & 80.01\%  &  81.09\%     &  72.36\% & 72.22\%  \\
     w/o span-whole\_word   & 79.46\%  &  80.70\%     &  70.73\% & 71.01\%   \\
     w/o visual pre-train tasks       & 79.73\%  &  80.89\%     &  71.04\% & 70.88\%  \\
     w/o acoustic pre-train task        & 79.48\%  &  80.82\%     &  71.78\% & 71.52\%  \\
    \hline
    \end{tabular}}
\caption{Ablation study of the pre-training tasks. ``span-whole\_word'' refers to the span masking strategy and whole work masking strategy. ``visual pre-train tasks'' refers to ``SpanMVFR'' and ``SpanMVFC-KL''. ``acoustic pre-train task'' refers to ``SpanMAFR''.}
\label{tab:ablation}
\vspace{-3pt}
\end{table}

\begin{figure}[htp]
\centering
\subfigure[IEMOCAP]{
\includegraphics[scale=0.50]{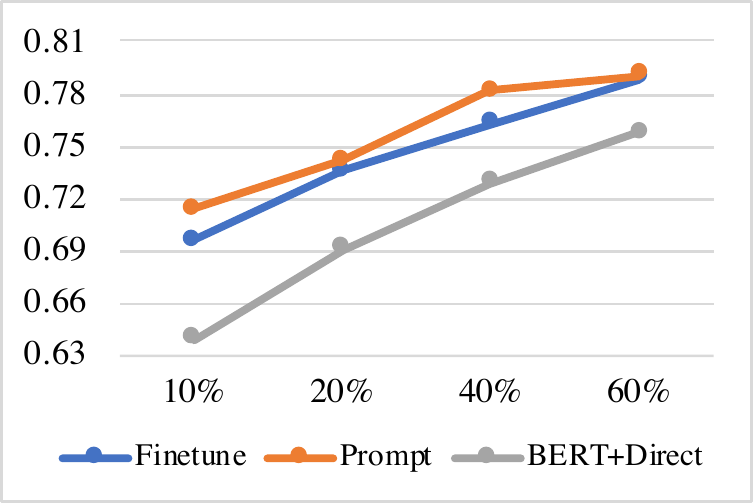}
}
\quad
\subfigure[MSP-IMPROV]{
\includegraphics[scale=0.5]{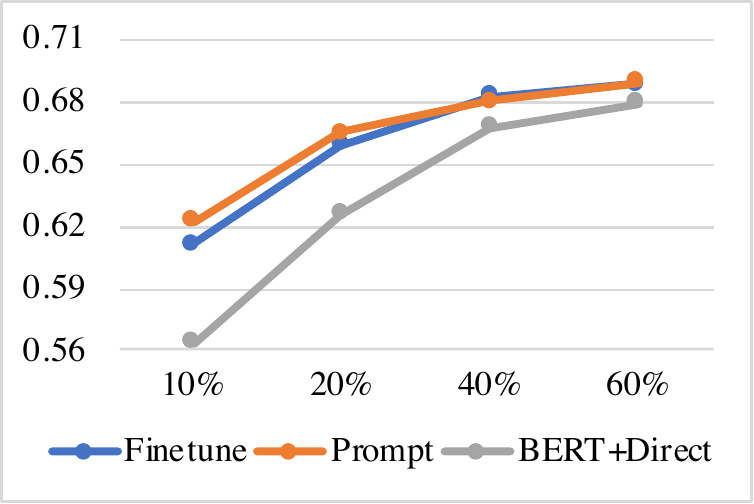}
}
\caption{Performance (UAR) comparison with different amount of training data. ``10\%'' means only 10\% of the training data for training.}
\label{fig:expr:fewshot}
\vspace{-5pt}
\end{figure}

\textbf{Ablation of different amounts of training data.}
In order to validate the generalization ability of the pre-trained MEmoBERT and prompt-based method under low-resource conditions, we conduct experiments using different amounts of training data in the downstream tasks.
As shown in Fig.~\ref{fig:expr:fewshot}, applying ``Finetune'' and ``Prompt'' on the pre-trained MEmoBERT both significantly outperforms the ``BERT+Direct'' setting under all low-resource conditions, and the less training data, the more obvious the improvement brought by the pre-trained MEmoBERT.
The prompt-based method outperforms the finetune-based method on IEMOCAP and MSP under almost all conditions. It indicates that the prompt-based method can better adapt the pre-trained MEmoBERT to downstream tasks, especially under low-resource conditions.

\section{Conclusion}
\label{sec:conclusion}
In this paper, we propose a novel multimodal transformer-based pre-trained model, MEmoBERT, under self-supervised learning on a large-scale unlabeled movie dataset for multimodal emotion recognition. We further investigate a prompt-based learning method that can better adapt the pre-trained MEmoBERT to downstream tasks. Extensive experiments on two public datasets demonstrate the effectiveness of our proposed methods.

\clearpage

\small
\bibliographystyle{IEEEbib}
\bibliography{refs}

\end{document}